\documentclass[letterpaper,10pt,conference]{ieeeconf}

\IEEEoverridecommandlockouts
\usepackage{amsmath,amssymb}
\usepackage{graphicx}
\usepackage{booktabs}
\usepackage{multirow}
\usepackage{xcolor}
\usepackage{url}
\usepackage[hidelinks]{hyperref}
\usepackage{subcaption}

\usepackage{stfloats}

\newif\ifreview
\reviewtrue

\newcommand{\method}{Res-HIL}

\newcommand{\vect}[1]{\boldsymbol{#1}}

\title{\method: Human-Guided Residual Reinforcement Learning\\
for Sample-Efficient Dexterous Manipulation}

\author{
Mariia Iavorskaia$^{1,2}$,
Christian Dietz$^{1}$,
Sebastian Albrecht$^{1}$,
and Majid Khadiv$^{2}$%
\thanks{$^{1}$Siemens AG, Research and Predevelopment, Germany.}%
\thanks{$^{2}$Technical University of Munich, Germany.}%
}

\begin{document}

\maketitle
\thispagestyle{empty}
\pagestyle{empty}

\begin{abstract}
Imitation learning enables robots to acquire manipulation skills from demonstrations, but the resulting policies can fail outside the training data, while collecting more demonstrations requires substantial human effort. Human-in-the-loop reinforcement learning uses corrective feedback during online training, but typically learns the complete task policy rather than refining a pretrained imitation policy. We introduce \method{}, a human-in-the-loop residual reinforcement learning framework that learns corrective actions on top of a frozen imitation policy. Each human intervention provides two complementary learning signals: direct supervision of the residual policy and reward shaping of preceding autonomous behavior. Res-HIL combines these signals with zero initialization of the residual policy to stabilize and accelerate online learning. We evaluate Res-HIL on five contact-rich manipulation tasks spanning high-precision and long-horizon behaviors. With only 20 initial demonstrations, Res-HIL outperforms state-of-the-art full-policy human-in-the-loop reinforcement learning and residual fine-tuning without human guidance on every task after ten minutes of online training. Res-HIL improves its pretrained base policies and outperforms imitation policies trained with five times more demonstrations. An ablation study shows that direct residual supervision is critical to performance, while intervention-aware reward shaping substantially improves training efficiency. Videos are available on the \href{https://iavorskaiamariia.github.io/Res-HIL-website/}{\textcolor{blue}{\underline{project website}}}.
\end{abstract}

\section{Introduction}
\label{sec:introduction}

Learning complex robotic manipulation skills from human demonstrations has become an effective approach for acquiring behaviors that are difficult to specify manually~\cite{osa2018algorithmic,zhao2023act}. However, policies trained with behavior cloning are limited by the coverage of the demonstration data and may encounter states poorly represented during training~\cite{ross2011reduction,mandlekar2021matterslearningofflinehuman}. Small prediction errors can therefore lead to distribution shift and compound over time, making recovery particularly challenging in contact-rich and long horizon manipulation tasks. Collecting additional demonstrations can mitigate these failures, but requires substantial human effort and does not explicitly target the failure modes encountered by the deployed policy. This motivates methods that can efficiently improve a pretrained manipulation policy through targeted real-world interaction.

Human-in-the-loop (HIL) reinforcement learning (RL) addresses this challenge by incorporating human feedback directly into online policy learning~\cite{wang2018interventionaidedreinforcementlearning,luo2025hilserl}. During training, a human can intervene when the autonomous policy produces undesirable behavior, preventing unproductive interactions and providing corrective experience in states where the policy is insufficient. Recent HIL-RL approaches have demonstrated that this feedback can enable efficient learning of precise, contact-rich manipulation skills in a sample-efficient manner~\cite{luo2025hilserl}. Nevertheless, with increasing complexity and horizon length of tasks, learning the complete task policy through online interaction can require substantial robot interaction and human supervision times~\cite{zhou2025spire}.

\begin{figure}
    \includegraphics[width=\linewidth]{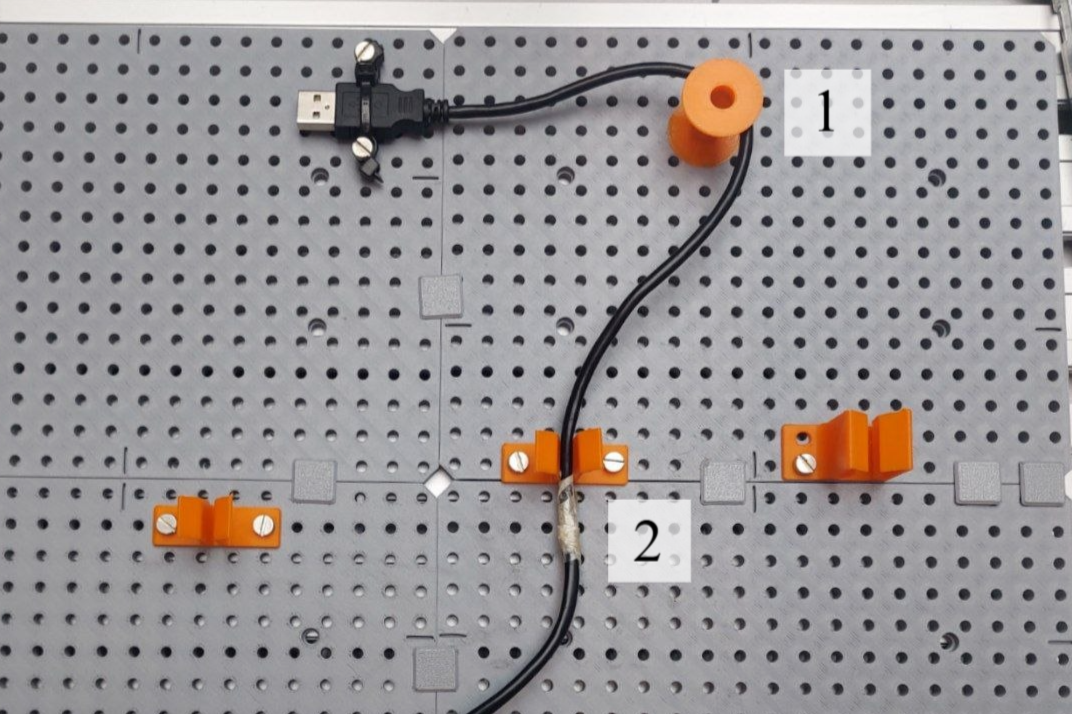}
    \caption{Illustration of the cable task. The robot must route the grasped cable around the cylinder (1) and insert its end into the socket (2), resulting in the successful final configuration shown on the image.}
    \label{fig:cable_image}
\end{figure}

Residual reinforcement learning offers an alternative by learning corrective actions on top of a policy previously trained from demonstrations~\cite{silver2019residual,ankile2025resfit}. This allows the agent to build on behavior captured by the base policy rather than relearning the complete task online. However, finding useful residual corrections still relies on autonomous exploration, which can be inefficient in real-world manipulation tasks where failures are difficult to recover from~\cite{eysenbach2017leavetracelearningreset}.

Motivated by these complementary strengths, we propose~\method{}, a human-in-the-loop residual reinforcement learning framework that uses human interventions not only for corrective control, but also as an additional learning signal for the residual policy. \method{} keeps the pretrained policy fixed and learns a residual policy to correct its actions during online interaction. Specifically, human-generated data provide direct supervision for the residual policy, while the occurrence of interventions is used to shape the otherwise sparse task reward. This allows \method{} to focus learning on the failures of the pretrained policy while building on behavior already captured by the base policy.

We evaluate \method{} on five real-world, contact-rich manipulation tasks spanning precise insertion and long-horizon cable manipulation, and compare it against imitation learning~\cite{zhao2023act}, HIL-SERL~\cite{luo2025hilserl}, and residual fine-tuning baselines~\cite{ankile2025resfit}. The results show that \method{} effectively improves pretrained policies and is particularly beneficial on challenging long-horizon tasks, where previously learned behavior provides a useful prior for online adaptation. Ablation experiments further show that direct supervision from human-generated data and intervention-aware reward shaping improve training and intervention efficiency.

The main contributions of this work are:
\begin{itemize}
    \item We introduce \method{}, a human-in-the-loop residual reinforcement learning framework for online adaptation of pretrained manipulation policies.

    \item We use each human intervention as two complementary learning signals: direct supervision for the residual policy and intervention-aware reward shaping.

    \item We evaluate \method{} on five real-world contact-rich manipulation tasks against imitation learning, HIL reinforcement learning, and residual fine-tuning baselines, and ablate its components.
\end{itemize}

\section{Related Work}
\label{sec:related_work}

\begin{figure*}[b]
    \centering
    \includegraphics[
        width=\textwidth,
    ]{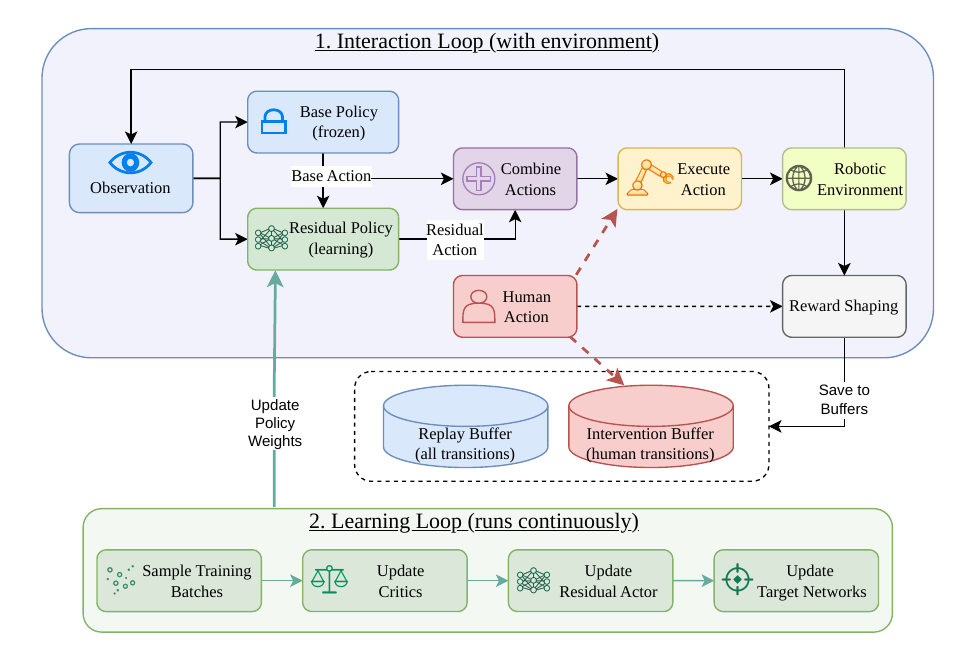}
    \caption{Overview of the proposed \method{} framework. During environment interaction, the frozen base policy and the learned residual policy jointly determine the executed action, while human interventions provide corrective actions and are incorporated into reward shaping. All transitions are stored in the replay buffer, with intervention transitions additionally stored in the intervention buffer. In parallel, the learning loop continuously samples training batches to update the critics and residual actor, followed by the corresponding target networks.}
    \label{fig:reshil_framework}
\end{figure*}

\subsection{Human-in-the-Loop Reinforcement Learning}

Human interventions have been incorporated into online reinforcement learning to replace unsafe or unproductive policy actions with expert control. Intervention-aided RL demonstrated this principle for real-world robot navigation~\cite{wang2018interventionaidedreinforcementlearning}. For manipulation, HIL-SERL~\cite{luo2025hilserl} combines initial demonstrations, online human interventions, and off-policy reinforcement learning to train visuomotor policies directly on physical robots. Recent systems extend this framework. ConRFT~\cite{chen2025conrftreinforcedfinetuningmethod} applies human-guided reinforcement learning to fine-tune pretrained vision-language-action policies, while SHaRe-RL~\cite{stranghöner2026sharerlstructuredinteractivereinforcement} incorporates task structure and compliance constraints for industrial assembly. E2HiL~\cite{deng2026e2hil} selects intervention samples according to their influence on policy entropy. PACT~\cite{liu2026preferencecalibratedhumanintheloopreinforcementlearning} uses intervention-induced preferences to correct value estimates for suboptimal trajectory segments, whereas OHP-RL~\cite{mo2026ohprlonlinehumanpreference} treats interventions as preference information and uses a state-dependent gate to regulate their influence on policy learning. In contrast to these methods, which optimize the complete task policy, Res-HIL keeps a pretrained imitation policy fixed and restricts online learning to residual corrections.

Intervention occurrence has also been used as a learning signal. Sirius~\cite{liu2023robotlearningjobhumanintheloop} labels a fixed-length window of autonomous transitions preceding each intervention as suboptimal and assigns these samples zero weight in its behavior cloning objective. PACT~\cite{liu2026preferencecalibratedhumanintheloopreinforcementlearning} uses a demonstration-trained progress model to localize suboptimal trajectory segments and an intervention-induced preference signal to correct their Bellman targets and guide actor optimization. In contrast, Res-HIL uses intervention timing to shape the reinforcement learning reward, while human-generated actions provide direct supervision for the residual policy.

\subsection{Residual Reinforcement Learning}

Residual reinforcement learning combines a fixed base controller with a learned policy that produces corrective actions. Johannink et al.~\cite{johannink2019residual} demonstrated this approach on real-world contact-rich assembly by learning residual corrections to a conventional feedback controller. Silver et al.~\cite{silver2019residual} further showed that residual policies can improve hand-designed and model-predictive controllers on long-horizon manipulation tasks.

Subsequent work applies residual learning to policies acquired from demonstrations. Alakuijala et al.~\cite{alakuijala2021residualreinforcementlearningdemonstrations} learn a visual base policy through behavior cloning and improve it with residual RL under sparse rewards. ResiP~\cite{ankile2024imitationrefinementresidual} augments a frozen action-chunked behavior cloning policy with a closed-loop residual policy for precise manipulation. Policy Decorator~\cite{yuan2024policydecoratormodelagnosticonline} introduces bounded residual actions and progressive exploration for model-agnostic refinement of large imitation policies, while ResFiT~\cite{ankile2025resfit} develops an off-policy residual learning procedure that reuses demonstration data and scales to high-DoF real-world manipulation. These methods establish residual learning as an effective means to improve fixed imitation policies, but rely primarily on autonomous exploration to discover useful corrections. Res-HIL instead uses targeted human interventions to directly identify corrective residual actions in states where the base policy fails.

\subsection{Human-in-the-Loop Residual Reinforcement Learning}
\label{sec:related_work:hil_res_rl}

Recent work has begun combining residual reinforcement learning with online human guidance. Focus-Then-Contact (FTC)~\cite{qiao2026focusthencontact} augments a frozen imitation policy with a residual policy, a keyframe-based visual affordance reward that guides exploration toward task-relevant contact regions, and a timed intervention mechanism that avoids conflicts between human and policy control. In contrast, Res-HIL does not require an auxiliary visual reward model. Instead, it derives additional learning signals directly from interventions: human actions supervise the residual policy, while intervention onset is used to penalize preceding autonomous transitions.

HiL-ResRL~\cite{liu2026hilresrlmodelagnosticfinetuningadapter} introduces a model-agnostic residual adapter for frozen visuomotor and vision-language-action policies and incorporates force--torque observations for contact-rich tasks. Human residual corrections replace the learned residual during intervention, and the resulting transitions are stored as demonstrations and online experience for mixed off-policy SAC updates. Res-HIL instead receives a full human action and explicitly constructs the residual target relative to the frozen base policy. This target directly supervises the residual actor through a behavior cloning objective, while intervention timing additionally shapes the rewards of preceding autonomous transitions. Thus, Res-HIL uses each intervention as both an action-level correction and a temporal indication of emerging failure.

\section{Method}
\label{sec:method}

\subsection{Problem Formulation}
We model the task as a discounted Markov decision process
$\mathcal{M}=(\mathcal{S},\mathcal{A},p,r,\gamma)$, where
$s_t\in\mathcal{S}$ is the environment state,
$\vect{a}_t\in\mathcal{A}$ is a continuous action,
$p(s_{t+1}\mid s_t,\vect{a}_t)$ denotes the transition dynamics,
$r_t=r(s_t,\vect{a}_t)$ is the sparse task reward, and $\gamma$
is the discount factor. The policy receives an observation $o_t$
consisting of camera images and proprioceptive measurements. A base policy $\pi_{\mathrm{BC}}$ is pretrained from demonstrations and remains frozen during online learning. The residual actor $\pi_{\theta}$, with $\theta$ being its parameters, produces
\begin{equation}
    \vect{a}^{\mathrm{res}}_t = \pi_{\theta}
    (o_t,\vect{a}^{\mathrm{base}}_t), \qquad
    \vect{a}^{\mathrm{base}}_t = \pi_{\mathrm{BC}}(o_t).
\end{equation}
The command sent to the robot is
\begin{equation}
    \vect{a}_t = \vect{a}^{\mathrm{base}}_t + \vect{a}^{\mathrm{res}}_t.
    \label{eq:combined_action}
\end{equation}

During online learning, the residual actor is optimized using Twin Delayed Deep Deterministic Policy Gradient (TD3)~\cite{fujimoto2018addressing}, while the pretrained base policy remains frozen.

\subsection{Human Interventions}
\label{sec:method:human}
At any time, the operator may override the policy with a full teleoperated action $\vect{a}^{\mathrm{human}}_t$. 

Policy and human transitions are stored in an online replay buffer $\mathcal{D}_{\mathrm{online}}$. Initial demonstrations and intervention transitions are stored in a demonstration buffer $\mathcal{D}_{\mathrm{offline}}$. Each training batch $\mathcal{B}$ contains equal numbers of samples from the two buffers, preventing rare human corrections from being underrepresented during optimization~\cite{ball2023efficient}.

The operator intervenes when the current motion is likely to cause failure,
damage, or an unrecoverable departure from the intended task progression. The
operator releases control once the system has returned to a recoverable state.
This rule allows the learner to experience correct recovery behavior without
requiring continuous teleoperation.

\subsection{Residual Policy Initialization}

To make the combined policy initially match the base policy, the output layer of the residual policy is initialized with zero weights and biases, with the resulting initial parameters denoted by $\theta_0$. Before any training, the residual policy predicts a zero correction for every observation,
\begin{equation}
    \vect{a}^{\mathrm{res}}_t = \pi_{\mathrm{\theta}_0}(o_t, \vect{a}^{\mathrm{base}}_t) = \mathbf{0},
\end{equation}
and the executed action is therefore
\begin{equation}
    \vect{a}_t = \vect{a}^{\mathrm{base}}_t + \vect{a}^{\mathrm{res}}_t
      = \vect{a}^{\mathrm{base}}_t.
\end{equation}
This initialization preserves the pretrained base policy’s behavior at the start of training. Only the residual policy’s output layer is initialized to zero; the preceding layers retain their standard random initialization, allowing the policy to learn corrective actions during training.

\subsection{Reward Shaping}

\begin{figure}[h]
    \centering

    \begin{subfigure}{\linewidth}
        \centering
        \includegraphics[width=\linewidth]{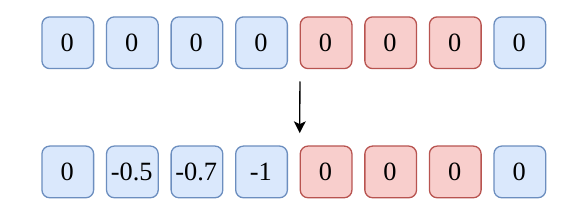}
        \caption{}
        \label{fig:intervention_penalty_a}
    \end{subfigure}

    \begin{subfigure}{\linewidth}
        \centering
        \includegraphics[width=\linewidth]{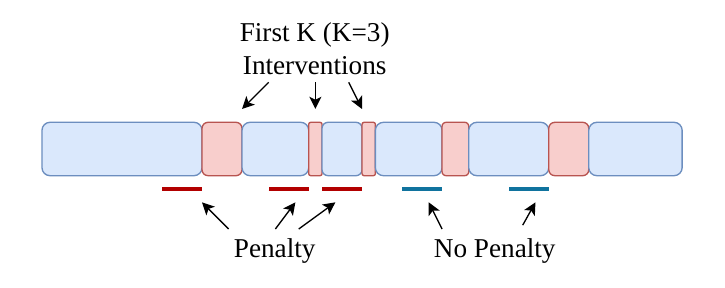}
        \caption{}
        \label{fig:intervention_penalty_b}
    \end{subfigure}

    \caption{Illustration of intervention-aware reward shaping. Blue blocks represent autonomous policy transitions, while red blocks represent human intervention transitions. (a) A penalty is assigned to the $I=3$ autonomous transitions immediately preceding an intervention, with its magnitude increasing as the intervention approaches. (b) Penalties are applied only to transitions preceding the first $K$ intervention segments of an episode, where $K = 3$ for illustration purposes; subsequent intervention segments do not trigger additional penalties.}
    \label{fig:intervention_penalty_comparison}
\end{figure}

The environment provides a sparse task reward $r_t$, which provides no intermediate signal before an intervention. We therefore use intervention onset as an additional learning signal by assigning a decaying penalty to the $I$ transitions preceding an intervention:
\begin{equation}
\tilde r_{t-i}=r_{t-i}-\lambda_{\mathrm{int}}\rho^{i},
\qquad i\in{0,\ldots,I-1},
\label{eq:intervention_penalty}
\end{equation}
where $\lambda_{\mathrm{int}}$ determines the magnitude of the intervention penalty and $\rho\in(0,1]$ controls its temporal decay. Thus, transitions immediately preceding an intervention receive the strongest penalty, while earlier transitions receive progressively weaker penalties, reflecting increasing uncertainty about their contribution to the intervention~(Fig. \ref{fig:intervention_penalty_a}). We choose $\lambda_{\mathrm{int}}$ relative to the scale of the task reward such that intervention-triggering behavior is penalized without overwhelming the sparse success signal.

The penalty is applied only before the first $K$ intervention segments of each episode~(Fig. \ref{fig:intervention_penalty_b}) to limit the cumulative shaping penalty in episodes with frequent interventions. This prevents intervention penalties from dominating the sparse task reward. A terminal penalty is additionally applied to failed episodes.

\subsection{Optimization Objective}

Following TD3, we learn two critic networks $Q_{\phi_1}(o_t,\vect{a}_t)$ and $Q_{\phi_2}(o_t,\vect{a}_t)$ that estimate the return of an executed action. For a transition $(o_t,\vect{a}_t,\tilde{r}_t,o_{t+1},d_t)$, where $d_t$ denotes the episode termination indicator, the TD3 Bellman target is
\begin{equation}
    y_t =
    \tilde{r}_t
    + \gamma(1-d_t)
    \min_{i\in\{1,2\}}
    Q_{\bar{\phi}_i}
    \left(o_{t+1},\vect{a}'_{t+1}\right),
\end{equation}
where the target action is obtained from the combined base and residual policies:
\begin{equation}
    \vect{a}'_{t+1}
    =
    \pi_{\mathrm{BC}}(o_{t+1})
    +
    \pi_{\bar{\theta}}
    \left(
        o_{t+1},
        \pi_{\mathrm{BC}}(o_{t+1})
    \right)
    + \vect{\epsilon},
\end{equation}
where $\bar{\phi}_i$ and $\bar{\theta}$ denote the target-network parameters, updated from the online networks by Polyak averaging with coefficient $\tau$, and $\vect{\epsilon}$ is the clipped target-policy smoothing noise used in TD3.

The critic loss is
\begin{equation}
    \mathcal{L}_{Q}=\sum_{i=1}^{2}
    \mathbb{E}_{\mathcal{B}}
    \left[
    \left(Q_{\phi_i}(o_t,\vect{a}_t)-y_t\right)^2
    \right].
\end{equation}

The residual actor is optimized with respect to the value of the combined action defined in Eq.~\eqref{eq:combined_action}. Following TD3~\cite{fujimoto2018addressing}, the actor objective is
\begin{equation}
    \mathcal{L}_{\mathrm{RL}}
    =
    -\mathbb{E}_{o_t\sim\mathcal{B}}
    \left[
        Q_{\phi_1}
        \left(
            o_t,\,
            \vect{a}^{\mathrm{base}}_t
            + \vect{a}^{\mathrm{res}}_t
        \right)
    \right].
    \label{eq:actor_rl_loss}
\end{equation}
To favor small corrections to the base policy, we regularize the magnitude of the residual action,
\begin{equation}
    \mathcal{L}_{\mathrm{res}}
    =
    \mathbb{E}_{o_t\sim\mathcal{B}}
    \left[
        \left\|
        \vect{a}^{\mathrm{res}}_t
        \right\|_2^2
    \right].
    \label{eq:residual_l2_loss}
\end{equation}

Following Fujimoto and Gu~\cite{fujimoto2021minimalist}, we additionally apply a behavior cloning objective to human corrections collected online. While TD3+BC applies the cloning objective to actions contained in a fixed offline dataset, we extend this objective to human corrections collected during online interaction.

For a transition corresponding to a demonstrated or intervened action
$\vect{a}^{\mathrm{human}}_t$, the target residual is defined relative to the frozen base policy as
\begin{equation}
    \vect{a}^{\mathrm{res},*}_t
    =
    \vect{a}^{\mathrm{human}}_t
    -
    \vect{a}^{\mathrm{base}}_t.
    \label{eq:target_residual}
\end{equation}
The behavior cloning loss is computed over the demonstration and intervention samples $\mathcal{B}_{\mathrm{demo/int}} \subseteq \mathcal{B}$:
\begin{equation}
    \mathcal{L}_{\mathrm{BC}}
    =
    \mathbb{E}_{(o_t,\vect{a}^{\mathrm{res},*}_t)
    \sim\mathcal{B}_{\mathrm{demo/int}}}
    \left[
        \left\|
        \vect{a}^{\mathrm{res}}_t
        -
        \vect{a}^{\mathrm{res},*}_t
        \right\|_2^2
    \right].
    \label{eq:bc_loss}
\end{equation}
Thus, the cloning objective preserves information from the initial demonstrations while continuously incorporating corrective actions provided through human interventions.

The resulting actor objective combines the TD3 objective, residual action regularization, and behavior cloning loss:
\begin{equation}
    \mathcal{L}_{\mathrm{actor}}
    =
    \mathcal{L}_{\mathrm{RL}}
    + \lambda_{\mathrm{res}}\mathcal{L}_{\mathrm{res}}
    + \lambda_{\mathrm{BC}}\mathcal{L}_{\mathrm{BC}},
    \label{eq:actor_loss}
\end{equation}
where $\lambda_{\mathrm{res}}$ and $\lambda_{\mathrm{BC}}$ control the contributions of the residual action penalty and behavior cloning objective, respectively.


\section{Experimental Evaluation}
\label{sec:experiments}


\subsection{Robot System and Tasks}
Experiments are conducted using a Universal Robots UR5 manipulator, a space mouse for teleoperation, two gripper-mounted cameras, and proprioceptive observations. Visual observations are encoded using ResNet-10~\cite{he2016deep} pretrained on ImageNet~\cite{russakovsky2015imagenet}. Actions specify relative Cartesian end-effector motion with respect to the current pose, including translational and rotational displacement.

\textbf{Peg-in-hole.} The robot must approach, align, and insert a peg under pose variation. In the easy variant, the peg orientation is randomized by up to $15^\circ$ about one axis relative to the gripper. In the hard variant, this range is increased to $25^\circ$, while the gripper orientation is additionally randomized by up to $10^\circ$ about each of two orthogonal axes. In both variants, the initial gripper position is randomized across episodes.

\textbf{Cable routing and insertion.} The robot must route a cable around an obstacle, then insert its end. Success requires completing both subtasks in sequence, making this the principal long-horizon evaluation. The cable grasp location and the initial gripper pose are randomized across episodes to introduce variation in the initial configuration. We consider two observation settings. In the two-camera setting, the policy receives observations only from the two wrist-mounted cameras and does not have a complete view of the workspace. In particular, the obstacle is not visible during cable routing. In the three-camera setting, an additional side camera provides a global view of the scene, allowing the policy to observe the obstacle throughout the cable-routing phase.

\textbf{Vent insertion.} The robot must insert a lid into an opening following a prescribed sequence: one edge must be inserted first, followed by the second edge. An episode is considered successful only if this insertion order is respected. Inserting the edges in the reverse order or pushing the lid directly through the opening is considered a failure, as such strategies could damage the components. The lid grasp and the initial gripper pose are randomized across episodes.

We used a fixed training budget of 20 minutes for the easy peg-in-hole task and 45 minutes for each of the remaining tasks.

\subsection{Baselines}
We compare the proposed Res-HIL method against the following baselines:
\begin{itemize}
    \item \textbf{ACT}: the frozen pretrained policy without online fine-tuning~\cite{zhao2023act};
    \item \textbf{ResFiT}: residual off-policy reinforcement learning without human interventions~\cite{ankile2025resfit};
    \item \textbf{HIL-SERL}: off-policy human-in-the-loop reinforcement learning that directly fine-tunes the full policy~\cite{luo2025hilserl}; and
    \item \textbf{Naive Res-HIL}: a baseline that directly combines residual reinforcement learning with human interventions, without the proposed residual initialization, intervention-aware reward shaping, or intervention-supervised BC objective.
\end{itemize}

All methods are initialized from the same set of 20 demonstrations, except ACT (100 demos) and ResFiT (100 demos), which use 100 demonstrations, and are evaluated on the same robotic platform. Online methods are compared in terms of both elapsed training time and collected environment transitions. Human-generated transitions are reported separately to quantify the amount of human supervision required during training.

The concurrent FTC~\cite{qiao2026focusthencontact} and HiL-ResRL~\cite{liu2026hilresrlmodelagnosticfinetuningadapter} methods, discussed in Sec.~\ref{sec:related_work:hil_res_rl}, became available shortly before submission and are therefore not included in the experimental comparison, as a controlled real-robot reproduction was not feasible within the available evaluation period.

\begin{table*}[t]
    \centering
    \caption{Success rates after 10 minutes of online training. All values are reported as percentages.}
    \label{tab:results_10min}
    \small
    \setlength{\tabcolsep}{8pt}
    \begin{tabular}{lccccc}
        \toprule
        \textbf{Method}
        & \textbf{Peg-in-Hole}
        & \textbf{Peg-in-Hole}
        & \textbf{Vent}
        & \textbf{Cable}
        & \textbf{Cable} \\
        & \textbf{Easy}
        & \textbf{Hard}
        &
        & \textbf{2 Cameras}
        & \textbf{3 Cameras} \\
        \midrule

        HIL-SERL
        & 90 & 30 & 10 & 0 & 0 \\

        ResFiT (100 demos)
        & 0 & 0 & 0 & 0 & 0 \\

        Naive Res-HIL
        & 12 & 0 & 0 & 0 & 0 \\

        \textbf{Res-HIL}
        & \textbf{100} & \textbf{64} & \textbf{50} & \textbf{66} & \textbf{92} \\

        \bottomrule
    \end{tabular}
\end{table*}

\begin{table*}[t]
    \centering
    \caption{Final success rates. ACT policies are evaluated without online fine-tuning. All values are reported as percentages.}
    \label{tab:results_final}
    \small
    \setlength{\tabcolsep}{8pt}
    \begin{tabular}{lccccc}
        \toprule
        \textbf{Method}
        & \textbf{Peg-in-Hole}
        & \textbf{Peg-in-Hole}
        & \textbf{Vent}
        & \textbf{Cable}
        & \textbf{Cable} \\
        & \textbf{Easy}
        & \textbf{Hard}
        &
        & \textbf{2 Cameras}
        & \textbf{3 Cameras} \\
        \midrule

        ACT (20 demos)
        & 36 & 28 & 44 & 46 & 70 \\

        ACT (100 demos)
        & 84 & 60 & 80 & 80 & 92 \\

        \midrule

        HIL-SERL
        & \textbf{100} & 94 & 84$^*$ & 0 & 0 \\

        ResFiT (100 demos)
        & 4 & 0 & 0 & 0 & 0 \\

        Naive Res-HIL
        & 24 & 12 & 12 & 0 & 0 \\

        \textbf{Res-HIL}
        & \textbf{100} & \textbf{96} & \textbf{100} & \textbf{88} & \textbf{100} \\

        \bottomrule
    \end{tabular}
    \vspace{2pt}
    \parbox{\textwidth}{%
        \vspace{4pt}
        \footnotesize
        $^*$HIL-SERL reaches 100\% if insertions in which the lid is pushed through the opening without following the prescribed insertion sequence are counted as successful.
    }
\end{table*}

\begin{table*}[t]
    \centering
    \caption{Average cycle time per successful episode across tasks in seconds. Lower values indicate faster task completion.}
    \label{tab:cycle_time}
    \small
    \setlength{\tabcolsep}{8pt}
    \begin{tabular}{lcccccc}
        \toprule
        \textbf{Method}
        & \textbf{Peg-in-Hole}
        & \textbf{Peg-in-Hole}
        & \textbf{Vent}
        & \textbf{Cable}
        & \textbf{Cable}
        & \textbf{Average} \\
        & \textbf{Easy}
        & \textbf{Hard}
        &
        & \textbf{2 Cameras}
        & \textbf{3 Cameras}
        & \\
        \midrule

        ACT (20 demos)
        & 6.37 & 8.17 & 6.79 & 7.40 & 9.90 & 7.73 \\

        ACT (100 demos)
        & 2.88 & 5.70 & 6.00 & \textbf{6.64} & \textbf{7.95} & 5.83 \\

        HIL-SERL
        & \textbf{2.21} & \textbf{2.65} & \textbf{2.04} & -- & -- & -- \\

        \textbf{Res-HIL}
        & 2.67 & 3.40 & 5.98 & 7.01 & 8.26 & \textbf{5.46} \\

        \bottomrule
    \end{tabular}
\end{table*}

\subsection{Evaluation Protocol}
Each online training experiment was repeated three times under the same conditions. Policy performance was evaluated at fixed intervals during training on 50 predefined randomized initial configurations for each task. The same evaluation configurations were used across methods and repetitions for consistent comparisons. Reported success rates are averaged over the repetitions. Across repetitions, the range of success rates at each evaluation point was at most 4 percentage points, while evaluations reaching 100\% success showed no variation. All HIL-based methods use the same interaction backend and control frequency and therefore collect environment transitions at the same rate. All human interventions across methods and repetitions were performed using the intervention criterion described in Sec.~\ref{sec:method:human} to ensure consistency in the intervention protocol.

Collecting 100 demonstrations required approximately 30 min for peg-in-hole, 45 min for vent insertion, and 60 min for cable manipulation, reflecting the increasing demonstration effort for longer-horizon tasks.

\section{Results}
\label{sec:results}

\subsection{Task Success}

Table~\ref{tab:results_10min} reports the success rates after 10 minutes of online training. \method{} achieves the highest success rate across all five tasks, reaching 100\% on the easy peg-in-hole task and 92\% on the three-camera cable task within this training budget. In contrast, HIL-SERL achieves lower success rates within the same budget, particularly on the more challenging tasks, while ResFiT and Naive Res-HIL achieve little or no success. These results demonstrate that \method{} can effectively leverage human interventions to rapidly improve the pretrained policy.

The final success rates are reported in Table~\ref{tab:results_final}. \method{} achieves 100\% success on three of the five tasks and at least 88\% success on every task. HIL-SERL eventually reaches comparable performance on the peg-in-hole and vent insertion tasks, but fails to solve either cable manipulation task. For vent insertion, HIL-SERL would reach 100\% if trials in which the lid was pushed through the opening without completing the intended insertion sequence were counted as successful. Under our stricter success criterion, which requires the prescribed edge-first insertion, these trials are considered failures, resulting in the reported 84\% success rate (marked with $^*$ in Table~\ref{tab:results_final}). ResFiT does not yield competitive performance despite being initialized with 100 demonstrations. In comparison, \method{} substantially improves over its 20-demonstration ACT base policy, from 46\% to 88\% on the two-camera cable task and from 70\% to 100\% on the three-camera variant.

Increasing the number of demonstrations from 20 to 100 substantially improves the frozen ACT policy, as shown in Table~\ref{tab:results_final}. Nevertheless, although initialized
with only 20 demonstrations, \method{} outperforms the 100-demonstration ACT policy across all five tasks through targeted online interaction and corrective human feedback.

Table~\ref{tab:cycle_time} compares the average cycle time of successful episodes. Among the methods evaluated across all five tasks, \method{} achieves the lowest overall cycle time, averaging 5.46\,s compared with 5.83\,s for ACT with 100 demonstrations and 7.73\,s for ACT with 20 demonstrations. Moreover, \method{} consistently reduces the cycle time relative to its 20-demonstration ACT base policy across all five tasks.

On the three insertion tasks, HIL-SERL produces the fastest successful trajectories, with cycle times between 2.04\,s and 2.65\,s. However, as shown in Table~\ref{tab:results_final}, HIL-SERL does not successfully solve either cable manipulation task, and therefore no corresponding cycle times are reported. In contrast, \method{} maintains low cycle times while achieving high success rates across all five tasks. Compared with its ACT base policy, the residual fine-tuning therefore improves both task success and the efficiency of successful executions.


\subsection{Training and Human Intervention Efficiency}

\begin{figure}
    \includegraphics[]{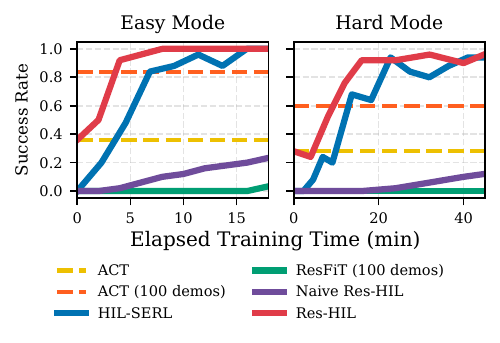}
    \caption{Success rate over elapsed training time for the easy and hard peg-in-hole tasks. Dashed horizontal lines indicate the performance of ACT policies trained with 20 and 100 demonstrations, while solid lines show performance during online training for HIL-SERL, ResFiT, Naive Res-HIL, and the proposed \method{} method. \method{} reaches high success rates earlier than the other online learning methods in both task variants.  }
    \label{fig:training_time_plots}
\end{figure}

Figure~\ref{fig:training_time_plots} compares success rates as a function of elapsed training time on the peg-in-hole tasks. On the easy variant, \method{} reaches over 90\% success within the first few minutes and converges to 100\% shortly thereafter. HIL-SERL reaches comparable performance later in training, whereas Naive Res-HIL remains below 25\%. A similar trend is observed on the hard variant, where \method{} reaches high success earlier than HIL-SERL, although both methods eventually achieve comparable final performance. These results indicate that the primary advantage of \method{} on these tasks lies in faster online adaptation rather than final performance alone.

\begin{figure}
    \includegraphics[]{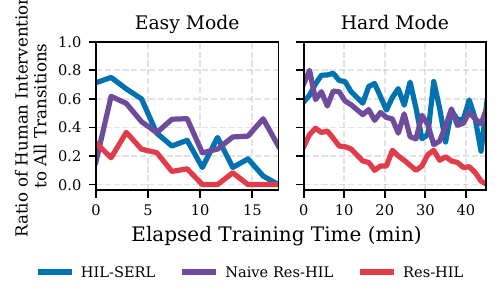}
    \caption{Proportion of human intervention transitions among
all collected transitions. Res-HIL achieves a high success rate while maintaining a consistently lower intervention ratio than HIL-SERL, indicating that it requires less human assistance over comparable training periods. A similar tendency was observed in the other task configurations.}
    \label{fig:interventions_ratio}
\end{figure}

Figure~\ref{fig:interventions_ratio} reports the ratio of human intervention transitions to all transitions collected during training. On the easy task, the intervention ratio of \method{} decreases as training progresses and approaches zero as the policy converges. On the hard task, human intervention remains necessary for longer; nevertheless, \method{} generally maintains a lower intervention ratio than HIL-SERL. Since human and autonomous transitions have equal duration and all HIL-based methods operate at the same control frequency, the intervention ratio directly corresponds to the fraction of interaction time under human control. Thus, the faster convergence observed in Fig.~\ref{fig:training_time_plots} is achieved without a higher proportion of human-generated transitions.

\subsection{Ablation Study}

We conduct an ablation study to evaluate the contribution of the individual components of \method{}. Specifically, we examine the effects of zero initialization, intervention-aware reward shaping, and the addition of a BC loss to the optimization objective. The experiments are conducted on the easy peg-in-hole task, which provides a controlled setting for isolating the effects of individual components on convergence and human intervention requirements while keeping the required real-world interaction tractable.

Removing zero initialization increases the convergence time from 8 to 16 minutes while retaining a final success rate of 100\%, with a moderately higher intervention ratio during early training. Removing intervention-aware reward shaping has a substantially larger effect: the policy does not converge within the 30-minute training budget and reaches 92\% success, while requiring persistent human intervention throughout training. The BC loss has the largest effect in this ablation; without it, the policy reaches only 20\% success within the training budget and maintains a high intervention ratio throughout most of training. These results show that zero initialization accelerates convergence, while reward shaping and, particularly, the BC loss are important for achieving efficient and reliable learning.

\begin{figure}[h]
    \centering
    \includegraphics[]{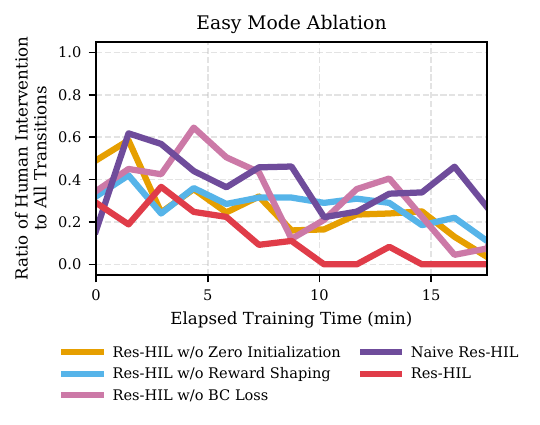}
    \caption{Human intervention ratio during training for component ablations on the easy peg-in-hole task.}
    \label{fig:ablation_studies}
\end{figure}

\begin{table}[h]
    \caption{Component ablation on easy Peg-in-Hole task.}
    \label{tab:ablations}
    \centering
    \small
    \begin{tabular}{lcc}
        \toprule
        Variant & Success (\%) & Time \\
        \midrule
        Full \method{} & 100 & 8m \\
        w/o zero initialization & 100 & 16m \\
        w/o reward shaping & 92 & 30m$^*$ \\
        w/o BC loss & 20 & 30m$^*$ \\
        \bottomrule
    \end{tabular}
    \parbox{\linewidth}{%
        \vspace{4pt}
        \footnotesize
        $^*$ Training was terminated after 30 minutes without reaching convergence.
    }
\end{table}

\section{Conclusion}
In this work, we introduced \method{}, a human-in-the-loop residual reinforcement learning framework for improving pretrained manipulation policies through online human feedback. Rather than relearning the complete behavior, \method{} learns residual corrections to a frozen base policy and uses human interventions both as corrective actions and as additional training signal. We evaluated the approach on five contact-rich manipulation tasks, ranging from precise insertion to long-horizon cable manipulation. The results show that combining a pretrained policy with human-in-the-loop residual learning is particularly beneficial on challenging long-horizon tasks, where the base policy alone is insufficient and learning the complete behavior online is difficult. On the peg-in-hole task, our ablations further show that intervention-aware reward shaping and additional supervision from human-generated data substantially improve training and human-intervention efficiency. 

The results suggest that pretrained policies and online human feedback provide complementary sources of information for efficient robot policy adaptation. Future work will investigate scaling \method{} to more diverse tasks and reducing the required human supervision.



\appendix
\section{Implementation Details}
\label{app:implementation}

\begin{table}[h]
    \caption{Hyperparameters used in all experiments}
    \label{tab:hyperparameters}
    \centering
    \small
    \begin{tabular}{ll}
        \toprule
        Parameter & Value \\
        \midrule
        Control frequency                    & $7\,\mathrm{Hz}$ \\
        Discount $\gamma$                    & $0.97$ \\
        Target update $\tau$                 & $0.005$ \\
        Actor learning rate                  & $3\times10^{-4}$ \\
        Critic learning rate                 & $3\times10^{-4}$ \\
        Batch size                           & $64$ \\
        Demo / online batch ratio            & $0.5$ / $0.5$ \\
        Policy delay                         & $2$ \\
        Target policy noise / clip           & $0.2$ / $0.5$ \\
        $\lambda_{\mathrm{int}}$             & $0.05$ \\
        $\lambda_{\mathrm{res}}$             & $0.005$ \\
        $\lambda_{\mathrm{BC}}$              & $0.4$ \\
        Gradient clipping (actor / critic)   & $1.0$ / $10.0$ \\
        Success / failure reward             & $10$ / $-2$ \\
        Penalized intervention segments $K$  & $2$ \\
        Penalized actions $I$                & $3$ \\
        Intervention decay $\rho$            & $0.6$ \\
        \bottomrule
    \end{tabular}
\end{table}

\bibliographystyle{IEEEtran}
\bibliography{references}

@article{luo2025hilserl,
    title={Precise and dexterous robotic manipulation via human-in-the-loop reinforcement learning},
    author={Luo, Jianlan and Xu, Charles and Wu, Jeffrey and Levine, Sergey},
    journal={Science Robotics},
    year={2025},
}

@misc{ankile2025resfit,
    title={Residual Off-Policy RL for Finetuning Behavior Cloning Policies}, 
    author={Lars Ankile and Zhenyu Jiang and Rocky Duan and Guanya Shi and Pieter Abbeel and Anusha Nagabandi},
    year={2025},
    archivePrefix={arXiv},
}

@article{fujimoto2021minimalist,
    title={A minimalist approach to offline reinforcement learning},
    author={Fujimoto, Scott and Gu, Shixiang Shane},
    journal={Advances in neural information processing systems},
    year={2021}
}

@inproceedings{zhao2023act,
    author    = {Zhao, Tony Z. and Kumar, Vikash and Levine, Sergey and Finn, Chelsea},
    title     = {Learning Fine-Grained Bimanual Manipulation with Low-Cost Hardware},
    booktitle = {Proceedings of Robotics: Science and Systems},
    year      = {2023}
}

@inproceedings{johannink2019residual,
  author    = {Johannink, Tobias and Bahl, Shikhar and Nair, Ashvin and Luo, Jianlan and Kumar, Avinash and Loskyll, Matthias and Ojea, Juan Aparicio and Solowjow, Eugen and Levine, Sergey},
  title     = {Residual Reinforcement Learning for Robot Control},
  booktitle = {Proceedings of the IEEE International Conference on Robotics and Automation},
  year      = {2019}
}

@inproceedings{zhou2025spire,
  author    = {Zhou, Zichen and Garg, Animesh and Fox, Dieter and Garrett, Caelan Reed and Mandlekar, Ajay},
  title     = {{SPIRE}: Synergistic Planning, Imitation, and Reinforcement Learning for Long-Horizon Manipulation},
  booktitle = {Proceedings of the Conference on Robot Learning},
  year      = {2024}
}

@inproceedings{stranghöner2026sharerlstructuredinteractivereinforcement,
  author    = {Strangh{\"o}ner, Jannick and Hartmann, Philipp and Braun, Marco and Wrede, Sebastian and Neumann, Klaus},
  title     = {{SHaRe-RL}: Structured, Interactive Reinforcement Learning for Contact-Rich Industrial Assembly Tasks},
  booktitle = {Proceedings of the IEEE International Conference on Robotics and Automation},
  year      = {2026}
}

@inproceedings{chen2025conrftreinforcedfinetuningmethod, 
    title={ConRFT: A Reinforced Fine-tuning Method for VLA Models via Consistency Policy}, 
    author={Yuhui Chen and Shuai Tian and Shugao Liu and Yingting Zhou and Haoran Li and Dongbin Zhao}, 
    booktitle={Proceedings of Robotics: Science and Systems},
    year={2025},
}

@inproceedings{qiao2026focusthencontact,
    title={Focus-Then-Contact: Speeding Up Robotic Contact-Rich Task Learning with Affordance-Guided Real-World Residual Reinforcement Learning},
    author={Guanren Qiao and Ruixiang Ouyang and Sheng Xu and Ruixing Jin and Yueci Deng and Yunxin Tai and Kui Jia and Guiliang Liu},
    booktitle={Forty-third International Conference on Machine Learning},
    year={2026},
}

@misc{silver2019residual,
      title={Residual Policy Learning}, 
      author={Tom Silver and Kelsey Allen and Josh Tenenbaum and Leslie Kaelbling},
      year={2019},
      eprint={1812.06298},
      archivePrefix={arXiv},
      primaryClass={cs.RO},
}

@article{osa2018algorithmic,
  author  = {Takayuki Osa and Joni Pajarinen and Gerhard Neumann
             and J. Andrew Bagnell and Pieter Abbeel and Jan Peters},
  title   = {An Algorithmic Perspective on Imitation Learning},
  journal = {Foundations and Trends in Robotics},
  year    = {2018},
}

@inproceedings{ross2011reduction,
  author    = {St{\'e}phane Ross and Geoffrey J. Gordon
               and J. Andrew Bagnell},
  title     = {A Reduction of Imitation Learning and Structured
               Prediction to No-Regret Online Learning},
  booktitle = {Proceedings of the Fourteenth International Conference
               on Artificial Intelligence and Statistics},
  year      = {2011}
}

@inproceedings{mandlekar2021matterslearningofflinehuman,
  author    = {Mandlekar, Ajay and Xu, Danfei and Wong, Josiah and Nasiriany, Soroush and Wang, Chen and Kulkarni, Rohun and Fei-Fei, Li and Savarese, Silvio and Zhu, Yuke and Mart{\'i}n-Mart{\'i}n, Roberto},
  title     = {What Matters in Learning from Offline Human Demonstrations for Robot Manipulation},
  booktitle = {Proceedings of the 5th Conference on Robot Learning},
  year      = {2021}
}

@inproceedings{wang2018interventionaidedreinforcementlearning,
  title={Intervention aided reinforcement learning for safe and practical policy optimization in navigation},
  author={Wang, Fan and Zhou, Bo and Chen, Ke and Fan, Tingxiang and Zhang, Xi and Li, Jiangyong and Tian, Hao and Pan, Jia},
  booktitle={Conference on Robot Learning},
  year={2018},
}

@inproceedings{
  eysenbach2017leavetracelearningreset,
  title={Leave no Trace: Learning to Reset for Safe and Autonomous Reinforcement Learning},
  author={Benjamin Eysenbach and Shixiang Gu and Julian Ibarz and Sergey Levine},
  booktitle={International Conference on Learning Representations},
  year={2018},
}

@article{deng2026e2hil,
   title={E2HiL: Entropy-Guided Sample Selection for Efficient Real-World Human-in-the-Loop Reinforcement Learning},
   journal={IEEE Robotics and Automation Letters},
   author={Deng, Haoyuan and Lin, Yudong and Xue, Yuanjiang and Du, Haoyang and Wang, Qianzhun and Zhou, Boyang and Wu, Zhenyu and Wang, Ziwei},
   year={2026},
}

@misc{liu2026preferencecalibratedhumanintheloopreinforcementlearning,
      title={Preference-Calibrated Human-in-the-Loop Reinforcement Learning for Robotic Manipulation}, 
      author={Zeyi Liu and Guangyao Liu and Yinuo Qu and Yuquan Xue and Bofang Jia and Chunhua Yang and Weihua Gui and Keke Huang and Ziwei Wang},
      year={2026},
      archivePrefix={arXiv},
}

@article{liu2023robotlearningjobhumanintheloop,
  title={Robot learning on the job: Human-in-the-loop autonomy and learning during deployment},
  author={Liu, Huihan and Nasiriany, Soroush and Zhang, Lance and Bao, Zhiyao and Zhu, Yuke},
  journal={The International Journal of Robotics Research},
  year={2025},
}

@misc{alakuijala2021residualreinforcementlearningdemonstrations,
      title={Residual Reinforcement Learning from Demonstrations}, 
      author={Minttu Alakuijala and Gabriel Dulac-Arnold and Julien Mairal and Jean Ponce and Cordelia Schmid},
      year={2021},
      archivePrefix={arXiv},
}

@inproceedings{ankile2024imitationrefinementresidual,
  title={From imitation to refinement -- residual rl for precise assembly},
  author={Ankile, Lars and Simeonov, Anthony and Shenfeld, Idan and Torne, Marcel and Agrawal, Pulkit},
  booktitle={Proceedings of the IEEE International Conference on Robotics and Automation},
  year={2025},
}

@inproceedings{yuan2024policydecoratormodelagnosticonline,
  title={Policy decorator: Model-agnostic online refinement for large policy model},
  author={Yuan, Xiu and Mu, Tongzhou and Tao, Stone and Fang, Yunhao and Zhang, Zhang and Su, Hao},
  booktitle={International Conference on Learning Representations},
  year={2025}
}

@misc{liu2026hilresrlmodelagnosticfinetuningadapter,
      title={HiL-ResRL: A Model-Agnostic Finetuning Adapter via Human-in-the-loop Residual Reinforcement Learning}, 
      author={Jingyi Liu and Zhaohong Mai and ShunSen He and Hang Ren and Chao Wang and Shunbo Zhou and XiaoDong Wu and Heng Zhang},
      year={2026},
      journal={arXiv preprint arXiv:2606.22860},
}

@inproceedings{fujimoto2018addressing,
  title={Addressing function approximation error in actor-critic methods},
  author={Fujimoto, Scott and Hoof, Herke and Meger, David},
  booktitle={International conference on machine learning},
  year={2018},
}

@misc{mo2026ohprlonlinehumanpreference,
      title={OHP-RL: Online Human Preference as Guidance in Reinforcement Learning for Robot Manipulation}, 
      author={Yunyang Mo and Jian Li and Qiwei Wu and Yihang Kang and Renjing Xu},
      year={2026},
}

@inproceedings{ball2023efficient,
  title={Efficient online reinforcement learning with offline data},
  author={Ball, Philip J and Smith, Laura and Kostrikov, Ilya and Levine, Sergey},
  booktitle={International Conference on Machine Learning},
  year={2023},
}

@inproceedings{he2016deep,
  author    = {Kaiming He and Xiangyu Zhang and Shaoqing Ren and Jian Sun},
  title     = {Deep Residual Learning for Image Recognition},
  booktitle = {Proceedings of the IEEE Conference on Computer Vision and Pattern Recognition},
  year      = {2016}
}

@article{russakovsky2015imagenet,
  title={Imagenet large scale visual recognition challenge},
  author={Russakovsky, Olga and Deng, Jia and Su, Hao and Krause, Jonathan and Satheesh, Sanjeev and Ma, Sean and Huang, Zhiheng and Karpathy, Andrej and Khosla, Aditya and Bernstein, Michael and others},
  journal={International journal of computer vision},
  year={2015},
}

\end{document}